\documentclass[10pt,twocolumn,letterpaper]{article}

\usepackage[pagenumbers]{wacv} % To force page numbers, e.g. for an arXiv version

\usepackage{graphicx}
\usepackage{amsmath}
\usepackage{amssymb}
\usepackage{booktabs}
\usepackage[accsupp]{axessibility} % Improves PDF readability for those with disabilities.
\usepackage[dvipsnames,table,xcdraw]{xcolor}

\newcommand{\red}[1]{\textcolor{BrickRed}{#1}}
\newcommand{\green}[1]{\textbf{\textcolor{ForestGreen}{#1}}}

\usepackage{xcolor}
\usepackage{enumitem}
\usepackage{url}
\usepackage{subcaption}
\usepackage{graphicx}
\usepackage{wrapfig}
\usepackage{multirow}
\usepackage{xspace}
\usepackage{interval}
\usepackage{colortbl}
\usepackage{pifont}
\usepackage{algorithm}
\usepackage{algpseudocode}
\usepackage{subcaption}
\def\argmax{\mathop{\rm argmax}}%
\definecolor{Gray1}{gray}{0.8}
\definecolor{Gray2}{gray}{0.9}
\usepackage[utf8]{inputenc}
\usepackage[T1]{fontenc}
\usepackage{listings}
\lstdefinestyle{pytorch}{
    language=Python,
    basicstyle=\small\ttfamily,
    keywordstyle=\color{blue},
    morekeywords={for, in, if, else},
    commentstyle=\color{green!40!black},
    numbers=left,
    numberstyle=\tiny,
    numbersep=5pt,
    frame=tb,
    columns=fullflexible,
    showstringspaces=false,
    captionpos=b,
    breaklines=true,
    escapeinside={(*@}{@*)},
}
\newcolumntype{C}{>{\centering\arraybackslash}p{2.5cm}}
\newcolumntype{L}{>{\arraybackslash}p{2.5cm}}
\usepackage[pagebackref,breaklinks,colorlinks]{hyperref}

\usepackage[capitalize]{cleveref}
\crefname{section}{Sec.}{Secs.}
\Crefname{section}{Section}{Sections}
\Crefname{table}{Table}{Tables}
\crefname{table}{Tab.}{Tabs.}

\def\wacvPaperID{903} % *** Enter the WACV Paper ID here
\def\confName{WACV}
\def\confYear{2025}

\begin{document}

\title{\textit{Just Shift It}: Test-Time Prototype Shifting for Zero-Shot Generalization with Vision-Language Models} 

\author{Elaine Sui \quad Xiaohan Wang \quad Serena Yeung-Levy \\
Stanford University \\
{\tt\small \{esui,xhanwang,syyeung@stanford.edu\}}}

\maketitle

\begin{abstract}
Advancements in vision-language models (VLMs) have propelled the field of computer vision, particularly in the zero-shot learning setting. Despite their promise, the effectiveness of these models often diminishes due to domain shifts in test environments. To address this, we introduce the \textbf{Test-Time Prototype Shifting} (\textbf{TPS}) framework, a pioneering approach designed to adapt VLMs to test datasets using unlabeled test inputs. Our method is based on the notion of modulating per-class prototypes in the shared embedding space. By pre-computing and caching prototypes generated with the pre-trained text encoder, TPS not only facilitates optimization-free prototype reuse for subsequent predictions but also enables seamless integration with current advancements in prompt engineering. At test-time, TPS dynamically learns shift vectors for each prototype based solely on the given test sample, effectively bridging the domain gap and enhancing classification accuracy. A notable aspect of our framework is its significantly reduced memory and computational demands when compared to conventional text-prompt tuning methods. Extensive evaluations across 15 image classification datasets involving natural distribution shifts and cross-dataset generalization, as well as in context-dependent visual reasoning, demonstrate TPS's superior performance, achieving state-of-the-art results while reducing resource requirements. Code is available at \url{https://github.com/elaine-sui/TPS}.
\end{abstract}
\section{Introduction}
\label{sec:intro}

\begin{figure}[t]
\begin{center}
\begin{subfigure}[t]{\linewidth}
    \includegraphics[width=\linewidth]{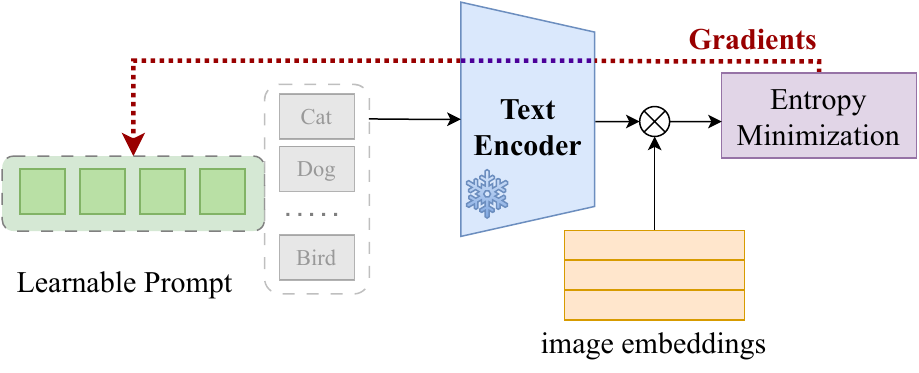}
   \caption{Test-Time Prompt Tuning}
   \label{fig:pull_figure_tpt}
\end{subfigure}
\hspace{1cm}
\begin{subfigure}[t]{\linewidth}
    \includegraphics[width=\linewidth]{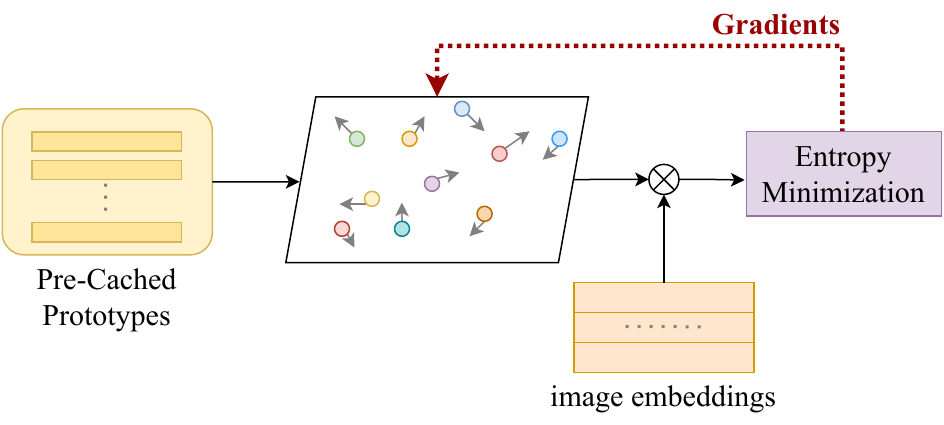}
   \caption{Test-Time Prototype Shifting}
   \label{fig:pull_figure_ours}
\end{subfigure}
\caption{Comparison of Test-Time Prompt Tuning (TPT)~\cite{shu2022tpt} against our method, Test-Time Prototype Shifting (TPS). TPT requires gradients to backpropagate through the large text encoder in order to reach the tuneable prompt, incurring high memory and computational costs. In contrast, TPS only backpropagates gradients to the feature space, in which our class prototype shifts are learned, making it much more efficient.}
\label{fig:pull_figure}
\end{center}
\vspace{-0.5cm}
\end{figure}

In recent years, the field of computer vision has witnessed remarkable progress, largely fueled by the emergence of robust vision-language foundation models~\cite{radford2021clip, girdhar2023imagebind}. These models, upon undergoing pre-training on large-scale datasets, acquire a deep understanding of visual concepts, enabling their seamless application to a range of downstream tasks without task-specific training.$_{\!}$ While these foundation models exhibit much better zero-shot generalization abilities compared to ImageNet pre-trained models, they still suffer from performance degradation due to domain shifts at test-time.

To address this problem, prior studies have explored various fine-tuning techniques, ranging from traditional full-model tuning to parameter-efficient methods~\cite{pmlr-v97-houlsby19a, gao2021clipadapter, lester-etal-2021-power, jia2022vpt}. Nevertheless, these strategies necessitate having enough labeled data for effective fine-tuning, posing challenges in domains where acquiring labeled data is difficult. In light of this, we turn our attention to the emerging paradigm of test-time adaptation (TTA), where model parameters are adjusted in an unsupervised manner with unlabeled test inputs.

How can we effectively enable test-time adaptation using these advanced vision-language models (VLMs)? There have been relatively few initiatives in this area. A notable example is the Test-Time Prompt Tuning (TPT) method~\cite{shu2022tpt}, which suggests fine-tuning several prompt tokens of the text input for each individual unlabeled test sample. However, this approach faces practical constraints due to the significant memory and computational demands involved in backpropagating through the text encoder for each image, as illustrated in Figure~\ref{fig:pull_figure}.

In this work, we propose \textbf{Test-Time Prototype Shifting} (\textbf{TPS}), a simple yet effective framework that specifically adjusts per-class prototypes within the embedding space. Initially, we compute each class prototype using the pre-trained text encoder from a VLM, enabling the prototypes to be cached and reused for all subsequent predictions. At test-time, we adapt by learning a shift vector for each prototype on the fly for a single test sample, bridging the domain gap between the prototypes and the target sample. A key highlight of our framework is that the shift vectors are the only parameters being updated and are adjusted within the embedding space itself, circumventing the need for backpropagation through large encoders. In comparison to TPT, our TPS framework achieves a \textbf{10x} increase in speed while necessitating less than \textbf{1/10} of the memory cost.

Benefiting from the two-stage learning paradigm, our approach fully capitalizes on the advancements in prompt engineering, such as using more sophisticated, well-designed prompts that significantly enhance generalization capabilities~\cite{radford2021clip, yang2023language, menon2022visual}, and using few-shot learned prompts~\cite{zhou2021coop, zhou2022cocoop}. In our study, we build upon these previous advancements by generating improved prototypes that can be seamlessly integrated into our Test-Time Prototype Shifting framework.

The zero-shot generalization capabilities of TPS were extensively evaluated across image classification and context-dependent visual reasoning. Image classification was evaluated with two distinct series of datasets: those involving natural distribution shifts and those focused on cross-dataset generalization. Our TPS consistently outperforms CLIP baselines, and surpasses current state-of-the-art (SoTA) by \textbf{3.3\%} on the natural distribution shifts benchmark and \textbf{1.9\%} on the cross-dataset generalization benchmark, respectively. Moreover, we demonstrate that regardless of the prototype-generation approach, learning a feature-space shift on the prototypes consistently boosts performance over zero-shot CLIP by over \textbf{4\%} on natural distribution shifts and up to \textbf{1\%} on cross-dataset generalization benchmarks. We further adapt TPS to context-dependent visual reasoning, outperforming current SoTA TTA methods on the Bongard-HOI benchmark~\cite{jiang2022bongard} by 1.3\%, on average. Remarkably, our approach not only outperformed TPT in terms of top-1 accuracy but also achieved this with only \textbf{1/10}-th of the memory and time costs. 

Our main contributions are summarized as follows:
\begin{enumerate}
 \item We introduce the Test-time Prototype Shifting (TPS) framework, a novel, straightforward and efficient approach. This is, to our knowledge, the first instance of utilizing feature space modulation for test-time adaptation with VLMs.
 \item Our TPS framework seamlessly integrates with existing advancements in prompt engineering, transforming it into a flexible plug-and-play system.
 \item We achieve state-of-the-art performance on image classification on both natural distribution shift and cross-dataset generalization benchmarks, surpassing current SoTA by 3.3\% and 1.9\%, respectively, as well as context-dependent visual reasoning, surpassing SoTA TTA methods by 1.3\%. Additionally, our approach significantly reduces computational and memory demands by more than 10 times compared to TPT.
\end{enumerate}
\section{Related Works}
\label{sec:related_works}

\subsection{Test-Time Adaptation}
Test-time adaptation (TTA) is the task of adapting a model's weights on an unlabeled out-of-distribution test set in order to achieve higher test performance.$_{\!}$ In the context of vision tasks, traditional methods leverage ImageNet-pretrained image classifiers and use techniques such as computing pseudo-prototype class representations to update the linear classifier~\cite{iwasawatest_time}, learning better feature representations through self-supervised auxiliary tasks~\cite{sun19ttt, liuttt+++, lin2023video}, adapting the normalization layers to learn the statistics of the target distribution~\cite{niu2023towards, niu2022efficient, schneider2020betterinc, song2023ecotta, wang2021tent}, as well as minimizing prediction entropy to increase the confidence of predictions~\cite{wang2021tent, niu2023towards, zhang2022memo, niu2022efficient, goyaltesttime}. With the development of CLIP~\cite{radford2021clip}, TTA methods have been predominantly based on prompt tuning. This involves learning a tunable text and/or image prompt to encode the visual distribution shift while maintaining the strong performances of these foundation models by keeping the pre-trained parameters fixed~\cite{shu2022tpt, hassan2023align, ma2023swapprompt, Feng_2023_ICCV, zhao2024testtime}. Despite only tuning a relatively small number of prompt parameters, tuning the input requires backpropagating through their respective encoders, which is especially memory intensive with large input sizes, making it infeasible in practice. Other works have proposed updating weights based on an auxiliary task~\cite{prabhudesai2023difftta}, but nevertheless, requires backpropagation through the model. In contrast, this work proposes to avoid backpropagation through the encoders and maintain the richness of the CLIP embedding space by directly modulating the features in it. 

Recent work have proposed training-free methods that directly modify the final predicted logit distribution. Some methods operate in an online streaming setting, using memory banks to store information on prior test inputs ~\cite{karmanov2024efficient,zhang2024dual,hu2024baftabackpropfreetesttimeadaptation}. This differs from our work where we assume that no prior knowledge of the test set is kept when predicting the label of each test example. Other training-free methods include adding a parameter-free attention module to modulate multi-modal features~\cite{guo2022calip} and computing the similarity between the target image and those from a constructed support set~\cite{udandarao2022sus-x}. Although our work does not directly fit this setting, we follow the same spirit of minimally adjusting intermediate representations to close the domain gap.

% \vspace{-3mm}
\subsection{Feature Modulation}
Feature modulation is a parameter-efficient tuning paradigm where features are perturbed to better conform to a target task. This learned perturbation is typically in the form of feature normalization, achieved by modulating the encoder's normalization layers to align source and target tasks~\cite{huang2017adain, Jaderberg_STN, perez2018film, cond_bn, chen2018on}. However, modulation can also be applied more directly to the features themselves~\cite{Lian_2022_SSF, zhou2023testtime}. For example, SSF~\cite{Lian_2022_SSF} proposes to learn scale and shift parameters for each layer's activations. DN~\cite{zhou2023testtime} proposes to subtract the means of the text and image embeddings from the respective inputs before computing CLIP similarity to align the CLIP training and inference procedures. We propose a more simplistic feature modulation procedure where we only learn shift vectors to pre-computed class prototypes in test-time training to better align them with the out-of-distribution image embeddings of the target dataset.

\subsection{Prompting for Vision-Language Models}
Vision-language models enable zero-shot generalization to downstream datasets via prompting. As predictions are computed by cosine similarity of the text and image embeddings, the quality of the text embeddings or class prototypes can cause a drastic difference in performance. In the case of image classification, this entails the careful design of natural language text descriptions for each of the class names, focusing on the visual aspects apparent from the image itself~\cite{radford2021clip}. CoOp~\cite{zhou2021coop} removes the need for hand-crafting prompts by prompt-tuning in the few-shot setting and CoCoOp~\cite{zhou2022cocoop} extends CoOp by learning instance-conditioned prompts, improving generalization ability. Another paradigm of prompt-engineering includes prompting large language models (LLMs) for better prompt templates~\cite{liu2023language} and/or content~\cite{menon2022visual, yang2023language}. Specifically, Menon and Vondrick~\cite{menon2022visual} and Yang \etal~\cite{yang2023language} both prompt GPT-3~\cite{gpt3} to generate concepts or descriptors of class names to increase zero-shot and linear-probe performance on image classification while providing model interpretability. Our work leverages these developments in prompt-engineering in our prototype generation phase, using these techniques to generate more knowledge-rich prototypes that can be swapped into the framework in a plug-and-play manner.
\section{Method}
\label{sec:method}

\begin{figure*}[t]
\begin{center}
\includegraphics[width=\linewidth]{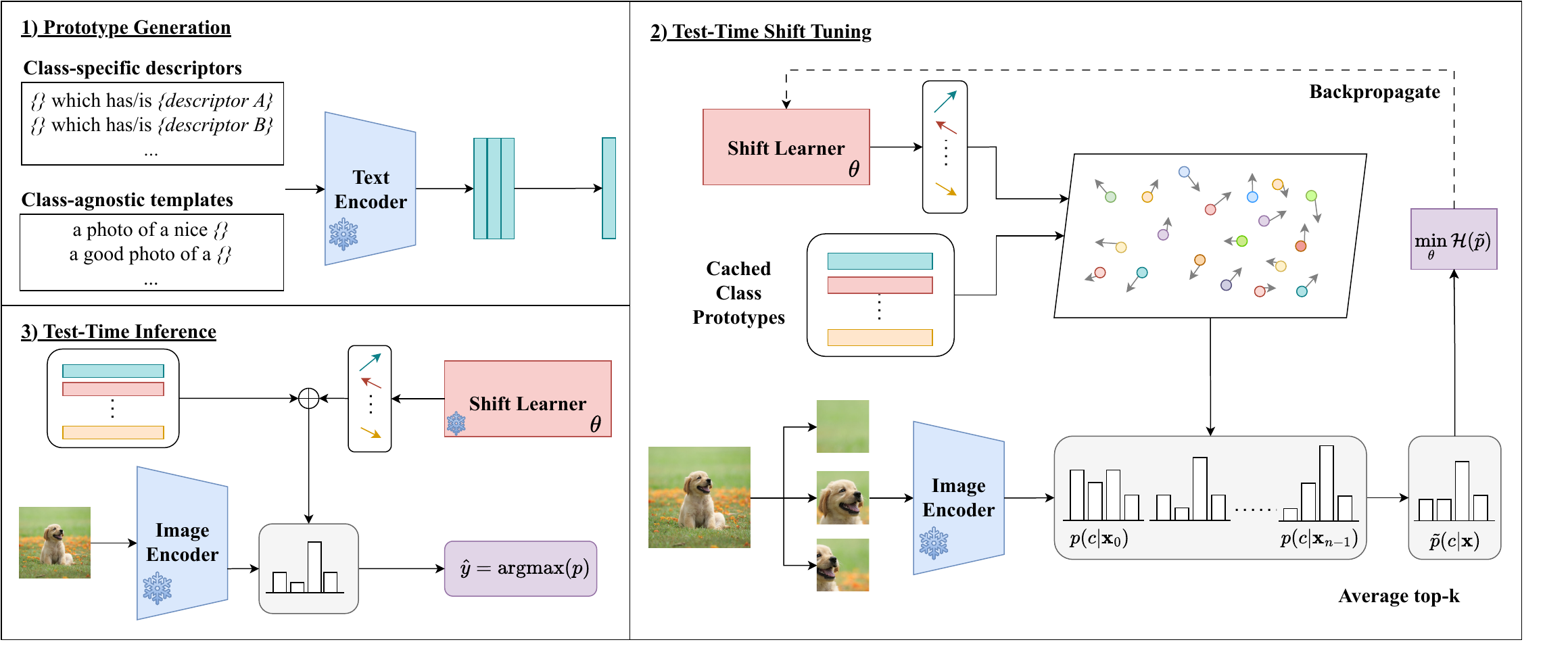}
   \caption{We illustrate the three stages of \textbf{Test-Time Prototype Shifting (TPS)}. \textbf{1) Prototype Generation:} pre-computation of class prototypes using different prompt-engineering strategies. We show the computation of $k$ class-conditioned descriptors for a single class. Means are computed and cached. \textbf{2) Test-Time Shift Tuning:} one iteration of test-time training where we tune the Shift Learner to generate small perturbations to the class prototypes to close the gap between the source and target distributions. Marginal entropy of the CLIP similarities of the shifted prototypes and augmented image embeddings is minimized. \textbf{3) Test-Time Inference:} Using the tuned Shift Learner, we compute the final prediction for the shifted class prototypes and the original image embedding with CLIP similarity.}
\label{fig:method}
\end{center}
\vspace{-0.5cm}
\end{figure*}

\subsection{Background}
\label{sec:background}
\paragraph{Test-Time Adaptation.}
Let $f_{\theta}$ be a model trained on data from some source distribution and $\mathcal{D}_{test}$ be some unlabeled out-of-distribution test dataset. Then, the goal of test-time adaptation is to adapt $\theta$ such that the performance of $f_{\theta'}$ on $D_{test}$ increases without the use of ground-truth test labels. Following \cite{shu2022tpt, memo}, we consider the single-input TTA setting where $f_{\theta}$ is adapted independently for every test example $d \in \mathcal{D}_{test}$. 

\paragraph{Image Classification with CLIP.}
CLIP~\cite{radford2021clip} is a large-scale vision-language model pre-trained by contrasting millions of (image, text) pairs. It consists of a text encoder $f_{\mathcal{T}}$ and an image encoder $f_{\mathcal{V}}$. To classify an image $v$ with CLIP, we use the CLIP pre-trained image encoder $f_{\mathcal{V}}$ to obtain a visual feature representation $\mathbf{x} = f_{\mathcal{V}}(v)$. Then, each label in the target label set $c \in \mathcal{C}_{target}$ is transformed with prompt template \texttt{p}. Each prompt is embedded with the pre-trained text encoder $f_{\mathcal{T}}$ to produce prototypes $\mathbf{p}_c = f_{\mathcal{T}}(\texttt{p}(c))$. The cosine similarities are computed between all class prototypes $\{\mathbf{p}_c\}_{c \in \mathcal{C}}$ and the visual embedding $\{\mathbf{x}\}$ to get a probability distribution over all classes.

\subsection{Test-Time Prototype Shifting}
\label{sec:method_overview}
Our method comprises three stages: Prototype Generation, Test-Time Shift Tuning, and Test-Time Inference, as depicted in Figure~\ref{fig:method}. 
Initially, in the Prototype Generation stage, we simply compute the class prototypes by embedding each class template $c \in \mathcal{C}$ of the test dataset. The vanilla prompt is ``\textit{a photo of a \{class\}.}'', with more advanced techniques discussed in Section~\ref{sec:advanced_prototype_generation}.
In the Test-Time Shift Tuning stage, shift vectors are generated through the Shift Learner to modify the prototypes (Section~\ref{sec:shift}), and cosine similarities between augmented image embeddings and shifted class prototypes are used to produce $n$ probability distributions. We optimize the Shift Learner to minimize the entropy of the aggregated marginal distribution. Finally, in the Test-Time Inference stage (Section~\ref{sec:overall_method}), we predict the class by comparing the cosine similarities between the learned shifted text embeddings and the original image embedding, choosing the class with the highest probability.

\subsubsection{Feature-Space Shift}
\label{sec:shift}
The Test-Time Prompt Tuning (TPT) method, as proposed by Shu \etal~\cite{shu2022tpt}, necessitates backpropagation through the text encoder at every test-time training step to learn the text prompt vectors. This process results in substantial computational and memory demands, which significantly hinders its practical applicability. 
Essentially, prompt tuning, when used for test-time adaptation, acts as an indirect technique for adjusting text embeddings. Its primary goal is to bridge the domain gap, simultaneously leveraging the multi-modal embedded knowledge of CLIP.
Such considerations bring us to a pivotal question: why not directly modulate class prototypes within the embedding space? 

This inspired the development of our Shift Learner module, designed specifically for learning modifications to the text embeddings. This approach not only capitalizes on the exceptional quality of the CLIP representation space but also adeptly avoids the requirement for the computationally demanding gradient backpropagation through the text encoder. Rather than indirectly altering embeddings through prompt tuning, our approach is to directly learn to shift class prototypes within the embedding space. This strategy enables us to preserve the overarching architecture of the CLIP embedding space, while adjusting the class prototypes for better alignment with out-of-distribution visual embeddings. 

We formally elaborate on the operation of our per-class shift as follows, given class $c \in \mathcal{C}$ and corresponding class prototype $\mathbf{p}_c \in \mathbb{R}^d$, we learn a shift vector $\mathbf{s}_c \in \mathbb{R}^d$.
The shift operation is performed by channel-wise addition between the class prototype $\mathbf{p}_c$ and the learned shift vector $\mathbf{s}_c$.
The normalized shifted prototype $\mathbf{p'}_c \in \mathbb{R}^d$ is generated as follows, 
\begin{align}
    \mathbf{p'}_c &= \frac{\mathbf{p}_c + \mathbf{s}_c}{||\mathbf{p}_c + \mathbf{s}_c||_2}
\end{align}

\subsubsection{Advanced Prototype Generation}
\label{sec:advanced_prototype_generation}
While it is effective to enhance the zero-shot generalization capabilities of vision-language models (VLMs) through test-time training, another line of techniques focuses on crafting sophisticated prompts to elevate performance. Our Test-Time Prototype Shifting framework is uniquely positioned in this landscape. By learning shifts on pre-computed and cached class prototypes, TPS is designed to be seamlessly compatible with any existing prompt-engineering methods. This integration empowers the generation of more robust and effective prototypes, leveraging advanced prompting strategies and offline prototype adjustment.

\textbf{Aggregating$_{\!}$ class-specific$_{\!}$ representations.}$_{\!}$  Our framework requires a single representation per class, but if such a representation is derived from a single prompt, the amount of information that it carries is limited to the number of input tokens that the CLIP text encoder was trained with. Hence, following~\cite{radford2021clip, menon2022visual}, we easily improve the robustness of class prototypes by taking the mean of the class-specific embeddings. This allows us to leverage multiple prompting and prompt-learning techniques and retain the knowledge from these various representations while keeping the computational and memory efficiency of our method without sacrificing prototype representation strength. We combine the class-agnostic prompts and per-class descriptors to generate the final prototype set $\{\mathbf{p}_c\}_{c \in \mathcal{C}}$. Several types of combinations are discussed in the Appendix.

\textbf{Prototypes as a plug-and-play module.} It is also important to note that our method is not limited to using these aforementioned prompting strategies and, provided that they are embedded within the CLIP text embedding space, is agnostic to how the prototypes are generated. In essence, this makes the prototypes a flexible plug-and-play module of our framework, enabling our framework to take advantage of future advancements in prototype creation.

\begin{algorithm}[t]
\caption{Test-Time Prototype Shifting (TPS)}
\begin{algorithmic}[1]
    \Require pre-trained and frozen image encoder $f_{\mathcal{V}}$ from a VLM, pre-computed class prototypes $\mathbf{p}_c$ and trainable shift parameters $\mathbf{s}_c$ for $c \in \mathcal{C}$, test image $v_0$, set of augmentations $\mathcal{A}$, number of augmentations $(n-1)$, optimizer \texttt{Opt}
    \\

    \Function{train}{$v_0$, $\mathcal{A}$, $f_\mathcal{V}$, $\{\mathbf{p}_c\}_{c \in \mathcal{C}}$, $\{\mathbf{s}_c\}_{c \in \mathcal{C}}$}
    \State \text{Sample} $a_1, \dots, a_{n-1} \in \mathcal{U}(\mathcal{A})$
    \State $v_i = a_i(v_0)$, $\mathbf{x}_i = f_{\mathcal{V}}(v_i)$ for $i \in \{0, \dots, n-1\}$
    \State $\mathbf{p}'_c = (\mathbf{p}_c + \mathbf{s}_c)/||\mathbf{p}_c + \mathbf{s}_c||_2$ for $c \in \mathcal{C}$
    \State Compute $p(c|\mathbf{x}_i, \mathbf{p}'_c)$ by Eq~\ref{eq:prob}
    \State Compute $\tilde{p} = \frac{1}{k}\sum_{i=1}^{k}p(c|\mathbf{x}'_{i}, \mathbf{p}'_c)$ for $\mathbf{x}'_{i}$ in top-$k$
    \State Compute $\mathcal{L}$ by Eq~\ref{eq:entropy} 
    \State Compute $\partial \mathcal{L}$
    \State Update $\{\mathbf{s}_c\}_{c \in \mathcal{C}}$: $\mathbf{s}_c \leftarrow \mathbf{s}_c - \texttt{Opt}(\partial \mathcal{L})$
    \EndFunction

    \\

    \Function{test}{$v_0$, $f_\mathcal{V}$, $\{\mathbf{p}_c\}_{c \in \mathcal{C}}$, $\{\mathbf{s}_c\}_{c \in \mathcal{C}}$}
    \State $\mathbf{x}_0 = f_{\mathcal{V}}(v_0)$
    \State $\mathbf{p}'_c = (\mathbf{p}_c + \mathbf{s}_c)/||\mathbf{p}_c + \mathbf{s}_c||_2$ for $c \in \mathcal{C}$
    \State Compute estimate $p(c|\mathbf{x}_0, \mathbf{p}'_c)$ by Eq~\ref{eq:prob}
    \State \Return $\argmax{p}$
    \EndFunction
\end{algorithmic}
\label{alg:method}
\end{algorithm}

\subsubsection{Test-Time Training and Inference}
\label{sec:overall_method}
At test time, given a single test image $v_0$, we follow~\cite{shu2022tpt} to augment it $(n-1)$ times and compute the features of the original and augmented images with the CLIP image encoder to obtain embeddings $\{\mathbf{x}_i\}_{i = 0}^{n-1}$, where $\mathbf{x}_0$ is the embedding of the original image $v_0$, and $\{\mathbf{x}_i\}_{i = 1}^{n-1}$ are the embeddings of the $(n-1)$ augmented versions of $v_0$.
As introduced in Section~\ref{sec:shift}, we shift the pre-cached prototypes $\{\mathbf{p}_c\}_{c \in \mathcal{C}}$ to obtain $\{\mathbf{p}'_c\}_{c \in \mathcal{C}}$. 
For each image feature $\mathbf{x}_i$, the predicted probabilities are calculated as
\begin{align}
\label{eq:prob}
    p(c| \mathbf{x}_i, \mathbf{p}'_c) &= \frac{\exp{(\mathbf{p'}_c^\top\mathbf{x}_i/\tau)}}{\sum_{c \in C} \exp{( \mathbf{p'}_c^\top\mathbf{x}_i}/\tau)}
\end{align}
where $\tau$ denotes the temperature scalar. 
Similar to TPT~\cite{shu2022tpt}, we select the $k$ distributions with highest confidence (\ie lowest entropy) of the batch to filter out uninformative and misleading views of the image, and compute the average distribution from the selected $k$ distributions. Denoting the image embeddings corresponding to the selected $k$ distributions as $\{\mathbf{x}'_i\}_{i = 1}^{k}$, we train our model to minimize the following entropy of this marginal distribution,
\begin{align}
\label{eq:entropy}
    \mathcal{L} &= -\sum_{c \in \mathcal{C}}\tilde{p}(c|\mathbf{x}_0, \mathbf{p}'_c)\log{\tilde{p}(c|\mathbf{x}_0, \mathbf{p}'_c)} \\
    \text{where } \tilde{p}(c|\mathbf{x}_0, \mathbf{p}'_c) &= \frac{1}{k}\sum_{i = 1}^{k}p(c|\mathbf{x}'_i, \mathbf{p}'_c)
\end{align}

This objective is used to encourage the model to make consistent, high-confidence predictions across multiple views, as model accuracy tends to correlate with model confidence~\cite{zhang2022memo,shu2022tpt}.

In our model, the only parameters that are optimized are the shift vectors $\{\mathbf{s}_c\}_{c \in \mathcal{C}}$. We update the shift vectors for a single step of gradient descent.

After test-time training, we encode the original image and compute its cosine similarity with the shifted class prototypes, resulting in a final prediction that is the \texttt{argmax} of the prediction logits. Algorithm~\ref{alg:method} summarizes the entire procedure of our proposed method, TPS, that enables efficient test-time adaptation using VLMs.

\subsubsection{TPS for Other Visual Tasks}
Though we demonstrate how our test-time prototype shifting (TPS) method can work for image classification, it can also be used in more complex tasks such as Context-Dependent Visual Reasoning. In this work, we adapt the method by Shu \etal~\cite{shu2022tpt} by simply substituting the prompt-tuning step with our proposed shift-tuning step. We refer the reader to this paper for more details of the method.
\section{Experimental Results}
\label{sec:experimental_results}

\subsection{Image Classification}
\paragraph{Datasets } We evaluate our method TPS on natural distributions shifts and cross-dataset generalization. For natural distribution shifts, we evaluate ImageNet~\cite{imagenet_cvpr09} along with its four variants: ImageNet-V2~\cite{imagenet_v2}, ImageNet-A~\cite{imagenet_a}, ImageNet-R~\cite{imagenet_r} and ImageNet-Sketch~\cite{imagenet_sketch}. For cross-dataset generalization, we evaluate on ten publicly available image classification datasets of different objects and scenes, including fine-grained and specialized datasets: Flowers102~\cite{flowers102}, DTD\cite{DTD}, OxfordPets~\cite{oxfordpets}, StanfordCars~\cite{stanfordcars}, UCF101~\cite{ucf101}, Caltech101~\cite{caltech101}, Food101~\cite{food101}, SUN397~\cite{sun397}, FGVC-Aircraft~\cite{aircrafts}, and EuroSAT~\cite{eurosat}. We report Top-1 accuracy for image classification on all datasets.

\paragraph{Implementation Details } Similar to TPT~\cite{shu2022tpt}, we augment a test image 63 times with random resized crops to obtain a batch of 64 images that also includes the original image. We select 10\% of samples in the batch with lowest entropy and compute the marginal entropy of the selected predicted probability distributions. We initialize the learnable shift to all zeros and optimize it for 1 step using the AdamW~\cite{loshchilov2018decoupled} optimizer and learning rate of 5e-3 for ImageNet variants and 1e-3 for cross-dataset generalization. For our method, we initialize each class prototype by taking the micro average of the mean of the class-agnostic CLIP template prompts and the mean of the class-specific GPT-4 generated descriptive prompts. 

\paragraph{Baselines } We compare our method with zero-shot and TTA baselines that leverage CLIP ViT-B/16 as a backbone. These TTA methods include TPT~\cite{shu2022tpt} which performs text-prompt tuning,  DiffTPT~\cite{Feng_2023_ICCV}, a variant of TPT that uses diffusion models to augment the visual training data, and Diffusion-TTA~\cite{prabhudesai2023difftta}, a method that adapts a discriminative classifier via a conditional diffusion model. To make our method more comparable to the simplest baselines, we also add their versions using more advanced prototypes, such as the inclusion of class-agnostic CLIP templates~\cite{radford2021clip} (\textbf{+~templates}), class-specific LLM-generated descriptors (\textbf{+~descriptors}), and learned CoOp~\cite{zhou2021coop} prompts (\textbf{+~CoOp}).

Given that Test-Time Prompt Tuning (TPT) does not involve tuning in the feature space, our advanced prompt generation is constrained to enhancing the input of TPT's text encoder. To address this, we append descriptors to the prompt suffixes, denoted as \textbf{+~descriptors*}. Moreover, a limitation arises with TPT's methodology of initializing its learnable prompt using a singular prompt template. This restricts the potential for integrating the diverse array of CLIP ImageNet prompt templates, as they are structured in various formats. As a result, to further augment the TPT-tuned class prototypes, we take its mean with the same advanced prototypes used at initialization in our method, denoted as \textbf{+(templates~+~descriptors)*}.

\subsubsection{Natural Distribution Shifts}
\label{sec:main_results_natural}

\begin{table*}[t]
  \centering
  \resizebox{0.98\linewidth}{!}{%
  \begin{tabular}{l|*{7}{C}}
    \toprule
    \textbf{Method}
    & \textbf{ImageNet} &  \textbf{ImageNet-A}  & \textbf{ImageNet-V2}       
    & \textbf{ImageNet-R} & \textbf{ImageNet-Sketch}  
    & \textbf{Average} & \textbf{OOD Average} \\
    \midrule

    & \multicolumn{7}{c}{\textit{\underline{Zero-Shot Baseline}}}  \\
    CLIP-ViT-B/16
    & {66.74} & {47.79} & {60.89}
    & {73.99} & {46.12} 
    & {59.10} & {57.20} \\
  
    \midrule
    & \multicolumn{7}{c}{\underline{\textit{Test-Time Adaption Baselines}}} \\
    
    TPT~\cite{shu2022tpt} 
    &  68.98  & 54.77  & 63.45      
    &  77.06  & 47.94 
    & {62.44} & 60.81   \\
    
    TPT~\cite{shu2022tpt}~+~descriptors*
    &  {67.71}  & {54.28}  & {61.24} 
    &  {73.39}  & {46.09}  
    & {60.54} & {58.75}  \\

    TPT~+~(templates~+~descriptors)*
    & {69.54} & {55.13} & {63.95}
    & {77.46} & {48.34} 
    & {62.88} & {61.22} \\  

    DiffTPT~\cite{Feng_2023_ICCV}
    &  {70.30}  & {55.68}  & {\textbf{65.10}} 
    &  {75.00}  & {46.80}  
    & {62.28} & {60.52}  \\

    Diffusion-TTA~\cite{prabhudesai2023difftta}
    & {63.8} & {-} & {-}
    & {-} & {-}
    & {-} & {-} \\

    \midrule
    \midrule
    
    \textbf{Ours} (Shift~+~templates~+~descriptors)
    & \cellcolor{Gray2}{\textbf{71.45}}\textcolor{blue}{$_{(\uparrow{4.71})}$}
    &  \cellcolor{Gray2}{\textbf{60.61}}\textcolor{blue}{$_{(\uparrow{12.82})}$}
    & \cellcolor{Gray2}{64.91}\textcolor{blue}{$_{(\uparrow{4.02})}$}
    & \cellcolor{Gray2}{\textbf{80.20}}\textcolor{blue}{$_{(\uparrow{6.21})}$}
    & \cellcolor{Gray2}{\textbf{50.88}}\textcolor{blue}{$_{(\uparrow{4.76})}$}
    & \cellcolor{Gray2}{\textbf{65.61}}\textcolor{blue}{$_{(\uparrow{6.51})}$} 
    & \cellcolor{Gray2}{\textbf{64.15}}\textcolor{blue}{$_{(\uparrow{6.95})}$}  \\

    \bottomrule
  \end{tabular}
}
  \caption{Acc@1 of zero-shot image classification with CLIP-ViT-B/16 backbone on ImageNet and its OOD variants. Performance improvements over zero-shot CLIP are denoted in \textcolor{blue}{(${\uparrow}$blue)}. Best performances are in \textbf{bold}. 
  }
    \label{tab:natural_dist_main}
\end{table*}
\begin{table*}[t]
  \centering
  \resizebox{0.98\linewidth}{!}{%
  \begin{tabular}{l|*{7}{C}}
    \toprule
    \textbf{Method}
    & \textbf{ImageNet} &  \textbf{ImageNet-A}  & \textbf{ImageNet-V2}       
    & \textbf{ImageNet-R} & \textbf{ImageNet-Sketch}  
    & \textbf{Average} & \textbf{OOD Average} \\
    \midrule

    & \multicolumn{7}{c}{\textit{\underline{Zero-Shot Baseline}}}  \\
    CLIP-ViT-B/16~+~CoOp~\cite{zhou2021coop}
    & {71.51} & {49.71} & {64.20}
    & {75.21} & {47.99} 
    & {61.72} & {59.28} \\

    \midrule
    & \multicolumn{7}{c}{\underline{\textit{Test-Time Adaption Baselines}}} \\

    TPT~+~CoOp~(\cite{shu2022tpt,zhou2021coop}) 
    & {73.61} & {57.95} & {66.83}
    & {77.27} & {49.29} 
    & {64.99} & {62.83} \\

    DiffTPT~+~CoOp~(\cite{Feng_2023_ICCV,zhou2021coop})    
    & {\textbf{75.00}}  & {58.09}  & {66.80}  
    & {73.90}  & {\textbf{49.50}}  
    & {64.12} & {61.97}  \\

    \midrule
    \midrule

    \textbf{Ours} (Shift~+~CoOp\cite{zhou2021coop})
    &  \cellcolor{Gray2}{73.73}\textcolor{blue}{$_{(\uparrow{2.22})}$}  & \cellcolor{Gray2}{\textbf{60.49}}\textcolor{blue}{$_{(\uparrow{10.78})}$}  & \cellcolor{Gray2}{\textbf{66.84}}\textcolor{blue}{$_{(\uparrow{2.64})}$} 
    &  \cellcolor{Gray2}{\textbf{77.44}}\textcolor{blue}{$_{(\uparrow{2.23})}$}  & \cellcolor{Gray2}{49.08}\textcolor{blue}{$_{(\uparrow{1.09})}$}  
    & \cellcolor{Gray2}{\textbf{65.52}}\textcolor{blue}{$_{(\uparrow{3.80})}$} & \cellcolor{Gray2}{\textbf{63.46}}\textcolor{blue}{$_{(\uparrow{4.18})}$}  \\

    \bottomrule
  \end{tabular}
}
  \caption{Acc@1 of zero-shot image classification with CLIP-ViT-B/16 backbone on ImageNet and its OOD variants using CoOp-learned prompts. Performance improvements over zero-shot CLIP are denoted in \textcolor{blue}{(${\uparrow}$blue)}. Best performances are in \textbf{bold}.
  }
    \label{tab:natural_dist_coop}
\end{table*}

Table~\ref{tab:natural_dist_main} presents the top-1 accuracy of our method, benchmarked against zero-shot and test-time adaptation (TTA) baselines using CLIP on ImageNet and its out-of-distribution variants. Our results demonstrate that shifting class prototypes significantly enhances performance. Compared to the baseline zero-shot CLIP, we observe an improvement of 7\%, and a 3.3\% increase over the vanilla TPT on average for out-of-distribution datasets. Furthermore, Table~\ref{tab:natural_dist_main} shows that directly appending descriptors to the TPT prompt suffixes results in a performance decrease of 2\%, emphasizing the limitations of TPT in seamlessly incorporating prompt-engineering techniques. Notably, Table~\ref{tab:natural_dist_coop} demonstrates that our approach of learning a feature-based shift outperforms TPT and DiffTPT by 0.6\% and 1.5\%, respectively, on average even when using advanced prototypes derived from learned CoOp~\cite{zhou2021coop} prompts without backpropagating through the text encoder or prompting a diffusion model. This finding underscores that feature space modulation can effectively replicate the impact of test-time prompt tuning in scenarios involving natural distribution shifts. 

\subsubsection{Cross-Dataset Generalization}
\label{sec:main_results_cross}

\begin{table*}[t]
  \centering
  
  \resizebox{\linewidth}{!}{%
  \begin{tabular}{l|*{11}{c}}
    \toprule
    \textbf{Method}
    & \textbf{Flower102} &  \textbf{DTD} & \textbf{Pets} & \textbf{Cars}
    & \textbf{UCF101}   & \textbf{CalTech101} & \textbf{Food101}   & \textbf{SUN397}
    & \textbf{Aircraft} & \textbf{EuroSAT} & \textbf{Average} \\
    \midrule

    & \multicolumn{11}{c}{\textit{\underline{Zero-Shot Baseline}}}  \\
    CLIP-ViT-B/16
    & {67.28} & {44.44} & {87.98} & {65.24}
    & {65.08} & {92.98} & {83.80} & {62.55}
    & {23.70} & {41.42} & {63.45} \\
  
    \midrule
    & \multicolumn{11}{c}{\underline{\textit{Test-Time Adaption Baselines}}} \\
    
    TPT~\cite{shu2022tpt} 
    & {68.98} & {47.75} & {87.79} & {66.87}
    & {68.04} & {94.16} & {84.67} & {65.50}
    & {24.78} & {42.44} & {65.10} \\

    TPT~\cite{shu2022tpt}~+~descriptors*
    & {69.14} & {\textbf{51.48}} & {86.05} & {64.84}
    & {70.10} & {93.59} & {81.83} & {65.44}
    & {22.29} & {42.98} & {64.77} \\

    TPT~+~(templates~+~descriptors)*
    & {69.71} & {46.93} 
    & {87.87} & {66.77}
    & {68.60} & {94.12} 
    & {84.94} & {66.11}
    & {23.37} & {43.17} 
    & {65.16}  \\

    DiffTPT~\cite{Feng_2023_ICCV}
    & {70.10} & {47.00} & {\textbf{88.22}} & {67.01}
    & {68.22} & {92.49} & {87.23} & {65.74}
    & {25.60} & {43.13} & {65.47} \\

    Diffusion-TTA~\cite{prabhudesai2023difftta}
    & {71.5} & {-}
    & {86.1} & {-}
    & {-} & {-}
    & {\textbf{88.8}} & {-}
    & {24.6} & {-}
    & {-} \\

    \midrule
    \midrule
    
    \textbf{Ours} (Shift~+~templates~+~descriptors)
    & \cellcolor{Gray2}{\textbf{71.54}}\textcolor{blue}{$_{(\uparrow{4.26})}$}
    &  \cellcolor{Gray2}{50.47}\textcolor{blue}{$_{(\uparrow{6.03})}$}
    & \cellcolor{Gray2}{87.35}\textcolor{blue}{$_{(\downarrow{0.63})}$}
    & \cellcolor{Gray2}{\textbf{69.06}}\textcolor{blue}{$_{(\uparrow{3.82})}$}
    & \cellcolor{Gray2}{\textbf{71.00}}\textcolor{blue}{$_{(\uparrow{5.92})}$}
    & \cellcolor{Gray2}{\textbf{95.09}}\textcolor{blue}{$_{(\uparrow{2.11})}$} 
    & \cellcolor{Gray2}{85.23}\textcolor{blue}{$_{(\uparrow{1.43})}$}
    &  \cellcolor{Gray2}{\textbf{68.98}}\textcolor{blue}{$_{(\uparrow{6.43})}$}
    & \cellcolor{Gray2}{\textbf{26.34}}\textcolor{blue}{$_{(\uparrow{2.64})}$}
    & \cellcolor{Gray2}{\textbf{44.48}}\textcolor{blue}{$_{(\uparrow{3.06})}$}
    & \cellcolor{Gray2}{\textbf{66.96}}\textcolor{blue}{$_{(\uparrow{3.51})}$}
    \\
    \bottomrule
  \end{tabular}
}
  \caption{Acc@1 of zero-shot image classification with CLIP-ViT-B/16 backbone on cross-dataset generalization. Performance improvements over zero-shot CLIP are denoted in \textcolor{blue}{(${\uparrow}$blue)}. Best performances are in \textbf{bold}.
  }
  \label{tab:cross_dataset}
\end{table*}

Table~\ref{tab:cross_dataset} details the performance of our approach relative to zero-shot and test-time adaptation (TTA) baselines when generalizing to fine-grained and specialized datasets. Our findings reveal that the optimal implementation of our method, which involves shifting class prototypes and integrating both CLIP templates and Large Language Model (LLM)-generated descriptors, results in an average improvement of 3.5\% over the zero-shot CLIP, and a 1.9\% increase compared to TPT. 

\subsection{Context-dependent Visual Reasoning}
\label{sec:main_results_bongard}

\paragraph{Dataset } We evaluate our method on the Bongard-HOI~\cite{jiang2022bongard} benchmark, where each test example consists of a small set of 6 positive support images $\mathcal{P}$ illustrating a specific human-object interaction (HOI) concept and 6 negative support images $\mathcal{N}$ that does not demonstrate that concept. The task is to identify whether a given query image $q$ belongs to the positive or negative set.

\paragraph{Implementation Details } 
We initialize the positive and negative class embeddings from a Gaussian distribution with $\sigma = 0.02$, and initialize the learnable shift embedding to all zeros. Similarly to Shu \etal~\cite{shu2022tpt}, we optimize the shift embedding to minimize the cross-entropy loss on the support set for 64 steps using the AdamW~\cite{loshchilov2018decoupled} optimizer and learning rate of 5e-3. Then, we classify the query image by cosine similarity of the shifted class embeddings and the query image embedding.

\paragraph{Baselines } We compare our method with fully-supervised and TTA baselines that leverage CLIP-ResNet50 
as a backbone. HOITrans~\cite{zou2021hoitrans} is a fully-supervised transformed-based detection method that matches HOIs in the query image to those in the support set. TPT~\cite{shu2022tpt} performs text-prompt tuning to encode the HOI concept in the prompt and learn optimal [CLS] embeddings.

\subsubsection{Human-Object Interaction}
\begin{table}[t]
\centering
\resizebox{\columnwidth}{!}{
\small
  \centering
  \resizebox{\linewidth}{!}{%
  \begin{tabular}{l|*{5}{c}}
    \toprule
    \textbf{Method} & \textbf{seen act.,} & \textbf{unseen act.,} & \textbf{seen act.,} & \textbf{unseen act.,} & \textbf{Average} \\
    & \textbf{seen obj.,} & \textbf{seen obj.,} & \textbf{unseen obj.,} & \textbf{unseen obj.,} & \\
    \midrule

    & \multicolumn{5}{c}{\underline{\textit{Fully-Supervised Baseline}}} \\
    HOITrans~\cite{zou2021hoitrans}
    &  {59.50}  & {64.38}  & {63.10} & {62.87} & {62.46} \\

    \midrule
    & \multicolumn{5}{c}{\underline{\textit{Test-Time Adaption Baseline}}} \\
  
    TPT~\cite{shu2022tpt}
    &  \textbf{{66.39}}  & {68.50}  & \textbf{{65.98}} & {65.48} & {66.59} \\

    \midrule
    \midrule

    \textbf{Ours} (Shift)
    & {66.30} & {\textbf{72.64}} & {65.63} & {\textbf{66.94}} & {\textbf{67.88}} \\
    \bottomrule
  \end{tabular}
}
}
  \caption{Acc on the Bongard-HOI benchmark with CLIP-ResNet-50 backbone. Best performances are in \textbf{bold}.}
  \label{tab:bongard_hoi}
\end{table}
Table~\ref{tab:bongard_hoi} demonstrates the accuracy of our method across the standard test splits of the Bongard-HOI benchmark. We demonstrate improved performance over existing fully-supervised baselines. Notably, we perform better than or on-par with TPT across all test splits, achieving a 1.28\% increase in performance, on average.

\subsection{Efficiency Analysis}
\label{sec:main_results_efficiency}

\begin{figure}[t]
\begin{center}
\includegraphics[width=0.9\linewidth]{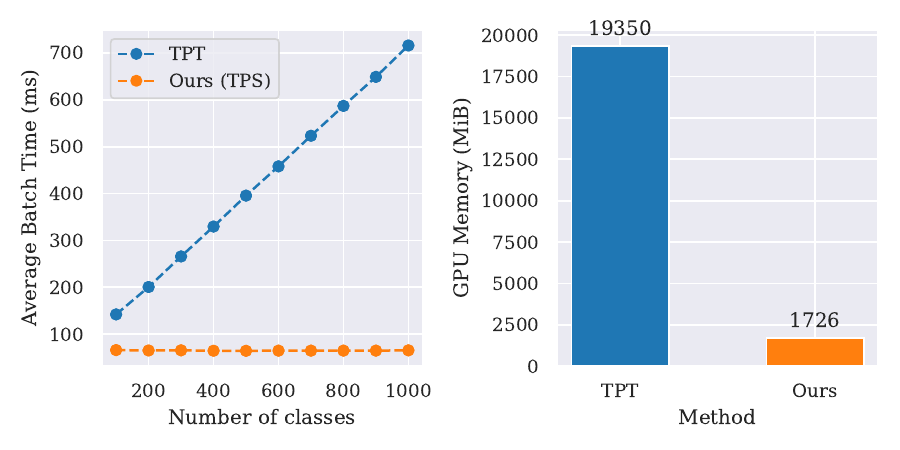}
\vspace{-1mm}
\caption{Comparison of computational and memory costs on an A6000 GPU on ImageNet. Left: Average runtimes of TPT and TPS across different sized subsets of ImageNet~\cite{imagenet_cvpr09} over 3 runs. Note that error bars are depicted but are not visible as they have extremely small standard deviations. Right: Memory consumption of TPT and TPS on ImageNet.}
\label{fig:efficiency}
\end{center}
\vspace{-5mm}
\end{figure}

The purpose of test-time adaptation is to tune a model on an out-of-distribution dataset at test-time given a single input or batch at a time. Given that the input is streaming, each adaptation requires low memory and computational cost to be used in practice. 
TPT~\cite{shu2022tpt}, by design, tunes the input features to the text encoder. Despite tuning few parameters overall, computing the gradient of the loss with respect to the prompt parameters necessitates the computation of gradients for the parameters in entire text encoder. Computing these gradients adds to the runtime, scaling linearly with the number of classes in the dataset, and retaining the encoder gradients adds to the memory requirements.

Figure~\ref{fig:efficiency} shows the average batch runtime and GPU memory consumption on ImageNet for TPS and the simplest TTA baseline TPT on a single A6000 GPU. We show the runtime of TPT and TPS on subsets of ImageNet comprised of different sized label sets. We observe that the average runtime per test input for TPT scales linearly with the size of the label set while our method, TPS, remains constant at approximately 65 ms per batch on average. On the full ImageNet test set, we further see that TPS runs \textbf{more than 10x as fast} and uses \textbf{less than 10x the memory} as TPT, and yet is able to achieve performance gains over the baseline. These significant speed-ups and low memory constraints enable our method to be easily used in practice.

\subsection{Ablation Studies}
\label{sec:ablations}

\subsubsection{Effect of Shift on Different Prototypes}
\label{sec:effect_of_shift}

\begin{table}[t]
  \centering
  \resizebox{\linewidth}{!}{
  \begin{tabular}{l|c|*{3}{c}}
    \toprule
    {\textbf{Prompt Type}} & {\textbf{Setting}}
    & {\textbf{ImageNet}} & {\textbf{ImageNet OOD}} & {\textbf{Cross-Dataset}} \\
    {} & {} & {} & \textbf{Average} & \textbf{Average} \\
    \midrule

    {Vanilla} & Zero-Shot
    & {66.74} & {57.20} & {63.45} \\

    & ~+~shift
    &  {68.77} & {61.59} & {64.41} \\

    \midrule
    & $\mathbf{\Delta}$ 
    &  \cellcolor{Gray2}{\green{{~+~2.03}}}  & \cellcolor{Gray2}{\green{{~+~4.39}}} & \cellcolor{Gray2}{\green{{~+~0.96}}}  \\

    \midrule
    \midrule
    
    {CoOp~\cite{zhou2021coop}} & Zero-Shot
    & {71.51} & {59.28} & {N/A} \\

    & ~+~shift
    & {73.73} & {63.46} & {N/A} \\

    \midrule
    & $\mathbf{\Delta}$
    &  \cellcolor{Gray2}{\green{{~+~2.22}}}  & \cellcolor{Gray2}{\green{{~+~4.18}}} & \cellcolor{Gray2}{N/A}  \\

    \midrule
    \midrule
    
    {CLIP templates} & Zero-Shot
    &  {68.35}  & {59.43} & {64.69}  \\

    & ~+~shift
    &  {70.38} & {64.04} & {65.57} \\

    \midrule
    & $\mathbf{\Delta}$
    &  \cellcolor{Gray2}{\green{{~+~2.03}}}  & \cellcolor{Gray2}{\green{{~+~4.61}}} & \cellcolor{Gray2}{\green{{~+~0.88}}}  \\

    \midrule
    \midrule

    {Descriptors} & Zero-Shot
    &  {68.52} & {58.29} & {66.02}  \\

    & ~+~shift
    &  {70.40} & {62.48} & {66.80} \\

    \midrule
    & $\mathbf{\Delta}$
    &  \cellcolor{Gray2}{\green{{~+~1.88}}} & \cellcolor{Gray2}{\green{{~+~4.19}}} & \cellcolor{Gray2}{\green{{~+~0.78}}}  \\

    \midrule
    \midrule
    
    {CLIP templates}
    & Zero-Shot
    &  {69.54} & {59.88} & {65.94} \\

    {~+~Descriptors}& ~+~shift
    &  {71.45} & {64.15} & {66.96} \\

    \midrule
    & $\mathbf{\Delta}$
    &  \cellcolor{Gray2}{\green{{~+~1.91}}} & \cellcolor{Gray2}{\green{{~+~4.27}}} & \cellcolor{Gray2}{\green{{~+~1.02}}}  \\

    \bottomrule
  \end{tabular}
}
  \caption{Acc@1 for zero-shot image classification comparing pure zero-shot vs. with learned feature-space shift from prototypes derived from various prompts using CLIP-ViT/B-16 backbone.
  }
  \label{tab:zero_shot_vs_shift}
\end{table}

We explore the effect of feature-space shift on a variety of prototypes. Specifically, we compare our method TPS against zero-shot performance given the same prototypes constructed from the vanilla \textit{``a photo of a \{class\}''} prompt, CoOp~\cite{zhou2021coop}-learned
 prompt, the 80 ImageNet context prompts from CLIP~\cite{radford2021clip} and our LLM-generated descriptors. As observed in Table~\ref{tab:zero_shot_vs_shift}, regardless of the prototype generation technique used, introducing even minor perturbations to class prototypes consistently yields an average gain of $>4$\% in top-1 accuracy on ImageNet out-of-distribution datasets and up to $1\%$ on cross-domain datasets over zero-shot CLIP with the same prototypes. This illustrates how the structure of the embedding space is maintained with a learnable shift.

\vspace{-0.2cm}
\subsubsection{Feature-Space Transformation Variants}
\label{sec:transformation_methods}

\begin{table}[t]
\centering
\resizebox{\columnwidth}{!}{
\small
  \centering
  \resizebox{\linewidth}{!}{%
  \begin{tabular}{l|*{3}{c}}
    \toprule
    {\textbf{Method}}
    & {\textbf{ImageNet}} & {\textbf{ImageNet OOD}} & {\textbf{Cross-Dataset}} \\
    & {} & {\textbf{Average}} & {\textbf{Average}}\\
    \midrule

    Scale
    &  {70.40}  & {61.12}  & {65.96} \\
  
    Shift
    &  \underline{71.45}  & \underline{64.15}  & \textbf{66.96} \\

    Scale~\&~Shift
    &  \textbf{71.47}  & \textbf{64.16}  & \underline{66.91} \\

    FiLM
    & {0.09} & {0.28} & {2.08} \\

    \bottomrule
  \end{tabular}
}
}
  \caption{Acc@1 of zero-shot image classification comparing different learned feature transformations using CLIP-ViT-B/16 backbone. Best performances are in \textbf{bold}. Second best performances are \underline{underlined}.}
  \label{tab:more-complex-transfo}
\end{table}

We compare different variants for feature-space transformations in Table~\ref{tab:more-complex-transfo}. Specifically, we compare against element-wise \textit{scale}, element-wise \textit{scale\&shift}, as well as \textit{FiLM}~\cite{perez2018film} where affine vectors are computed via a linear layer on the prototypes. We observe that \textit{scale} performs worse than \textit{shift}, and \textit{scale\&shift} performs equally to \textit{shift}. On the other hand, \textit{FiLM} suffers from model collapse due to much more learnable weights in TTA. Considering both efficiency and efficacy, \textit{shift} is the best choice in our framework.

\section{Conclusion}
\label{sec:conclusion}
We have presented the Test-time Prototype Shifting (TPS) framework, a novel approach to enhancing the zero-shot generalization abilities of VLMs. TPS addresses the limitations of existing test-time training methods by directly modulating class prototypes in the embedding space. This strategy not only reduces the computational and memory demands significantly but also allows for greater flexibility and precision in adapting to diverse domain shifts. By leveraging pre-computed and cached prototypes, and introducing class-specific shifts, TPS effectively bridges domain gaps. The extensive evaluations conducted across many datasets over two visual tasks demonstrate the superior performance of our method, outperforming existing approaches in terms of accuracy while being more computationally efficient.

\textbf{Acknowledgements}
This work was supported by the National Science Foundation grant No. 2026498.

%%%%%%%%% REFERENCES
{\small
\bibliographystyle{ieee_fullname}
\bibliography{egbib}
}

\appendix
\clearpage
\setcounter{page}{1}
\maketitlesupplementary

This document provides more details of our approach and additional experimental results, organized as follows:
\begin{itemize}
	\vspace{-3pt}
	\setlength{\itemsep}{0pt}
	\setlength{\parsep}{0pt}
	\setlength{\parskip}{0pt}
	\item \S~\ref{sec:s1} Implementation Details.
	\item \S~\ref{sec:s2} Additional Quantitative Results with Different Random Seeds.
	\item \S~\ref{sec:s3} Additional Ablation Studies.
        \item \S~\ref{sec:s4} Comparison to Training-Free Methods.
        \item \S~\ref{sec:s5} Research Impact and Limitations.
\end{itemize}

\section{Implementation Details of TPS}\label{sec:s1}
Algorithm~\ref{alg:pytorch_pseudocode} shows more detailed pseudocode in PyTorch-like style for Test-Time Prototype Shifting over an entire dataset. We will release the models and source code to ensure reproducibility.

\begin{algorithm*}[!htbp]
\caption{Test-Time Prototype Shifting Pseudocode in PyTorch-like style}
\label{alg:pytorch_pseudocode}
\begin{lstlisting}[style=pytorch]
# Define frozen parameters
image_encoder = CLIPImageEncoder()
prototypes = load_class_prototypes()

predictions = []
for img, label in data_loader:
    # Test-Time Shifting
    shift_params = nn.Parameter(torch.zeros(num_classes, embed_dim), requires_grad=True)
    aug_imgs = [aug(img) for i in range(batch_size - 1)]
    imgs = torch.stack([img] + aug_imgs, dim=0)
    image_features = image_encoder(imgs)

    text_features = prototypes + shift_params
    text_features = F.normalize(text_features, dim=-1)
    
    logits = (logit_scale * text_features @ image_features.T)

    # Confidence selection
    entropies = compute_batch_entropies(logits)
    top_k_idx = torch.argsort(batch_entropy, descending=False)[:k]
    
    loss = compute_average_entropy(logits[top_k_idx])
    optimizer.zero_grad()
    loss.backward()
    optimizer.step()

    # Test-Time Inference
    new_prototypes = prototypes + shift_params
    new_prototypes = F.normalize(new_prototypes, dim=-1)

    logits = (logit_scale * new_prototypes @ image_features[0].unsqueeze(0).T)
    pred = torch.argmax(logits)

    predictions.append(pred)

return predictions
    
    
\end{lstlisting}
\end{algorithm*}

\section{Main Results With More Random Seeds}\label{sec:s2}
In Sec~\ref{sec:natural_dist_more_seeds} and ~\ref{sec:cross_dataset_more_seeds}, we run Test-Time Prototype Shifting (TPS) over 3 random seeds on both the natural distribution shifts (Table~\ref{tab:natural_dist_main}) and cross-dataset generalization (Table~\ref{tab:cross_dataset}), respectively. The randomness comes from the image augmentation in creating a diverse minibatch for the entropy minimization objective. 

\subsection{Natural Distribution Shifts}
\label{sec:natural_dist_more_seeds}

\setcounter{table}{6}
\begin{table*}[t]
  \centering
  \resizebox{0.98\linewidth}{!}{%
  \begin{tabular}{l|*{7}C}
    \toprule
    \textbf{Method}
    & \textbf{ImageNet} &  \textbf{ImageNet-A}  & \textbf{ImageNet-V2}       
    & \textbf{ImageNet-R} & \textbf{ImageNet-Sketch}  
    & \textbf{Average} & OOD \textbf{Average} \\
    \midrule
    & \multicolumn{7}{c}{\textit{\underline{Test-Time Adaptation Baselines}}}  \\
    
    TPT~\cite{shu2022tpt} 
    &  {68.96} $(\pm {.03})$  & {54.47} $(\pm {.26})$  & {63.46} $(\pm {.07})$
    &  {77.10} $(\pm {.04})$  & {47.93} $(\pm {.03})$ 
    & {62.38} $(\pm {.05})$ & {60.74} $(\pm {.06})$  \\

    TPT~+~(templates~+~descriptors)*
    &  {69.51} $(\pm {.05})$  & {54.94} $(\pm {.17})$  & {63.86} $(\pm {.11})$
    &  {77.57} $(\pm {.11})$  & {48.38} $(\pm {.04})$ 
    & {62.85} $(\pm {.03})$ & {61.19} $(\pm {.04})$  \\    

    \midrule
    \midrule

    \textbf{Ours}
    &  {\textbf{71.43}} $(\pm {.06})$  & {\textbf{60.78}} $(\pm {.21})$  & {\textbf{65.00}} $(\pm {.09})$ 
    &  {\textbf{80.06}} $(\pm {.13})$  & {\textbf{50.97}} $(\pm {.09})$  
    & {\textbf{65.65}} $(\pm {.06})$ & {\textbf{64.20}} $(\pm {.08})$  \\

    \bottomrule
  \end{tabular}
}
  \caption{Acc@1 of zero-shot image classification with CLIP-ViT-B/16 backbone on ImageNet and its OOD variants over 3 random seeds. Best performances are in \textbf{bold}.
  }
    \label{tab:natural_dist_more_seeds}
\end{table*}

From Table~\ref{tab:natural_dist_more_seeds}, we observe that our conclusion from Sec~\ref{sec:main_results_natural} still holds. That is, our method outperforms SoTA TPT~\cite{shu2022tpt} by $> 3.4\%$ on average. We also observe that augmenting the TPT-tuned class prototypes with more advanced off-the-shelf prototypes only boosts performance by a mere $0.5\%$ on average over vanilla TPT, demonstrating TPT's limitation in maximally leveraging these advanced prototypes.

\subsection{Cross-Dataset Generalization}
\label{sec:cross_dataset_more_seeds}

\begin{table*}[t]
  \centering
  \resizebox{\linewidth}{!}{%
  \begin{tabular}{l|*{11}c}
    \toprule
    \textbf{Method}
    & \textbf{Flower102} &  \textbf{DTD} & \textbf{Pets} & \textbf{Cars}
    & \textbf{UCF101}   & \textbf{CalTech101} & \textbf{Food101}   & \textbf{SUN397}
    & \textbf{Aircraft} & \textbf{EuroSAT} & \textbf{Average} \\
    \midrule
    
    TPT~\cite{shu2022tpt} 
    & {68.79 $(\pm .1)$}  & {46.79 $(\pm .1)$}  
    & {87.09} $(\pm {.1})$ &  {66.38} $(\pm {.2})$ 
    & {67.86} $(\pm {.1})$  & {94.13} $(\pm {.1})$ 
    & {84.67} $(\pm {.1})$ & {65.41} $(\pm {.1})$ 
    & {23.44} $(\pm {.3})$ & {42.78} $(\pm {.3})$ 
    & {64.73} $(\pm {.1})$  \\

    TPT~+~(templates~+~descriptors)*
    & {69.67 $(\pm .11)$} & {47.56 $(\pm .55)$} 
    & {87.88 $(\pm .02)$} & {66.91 $(\pm .17)$}
    & {68.35 $(\pm .21)$} & {94.17 $(\pm .13)$} 
    & {84.89 $(\pm .07)$} & {66.23 $(\pm .12)$}
    & {23.55 $(\pm .31)$} & {43.12 $(\pm .18)$} 
    & {65.23 $(\pm .06)$}  \\

    \midrule
    \midrule

    \textbf{Ours}
    & {\textbf{71.47}} $(\pm {.12})$  & {\textbf{51.00}} $(\pm {.47})$  
    & {\textbf{87.45}} $(\pm {.09})$ &  {\textbf{68.99}} $(\pm {.10})$ 
    & {\textbf{70.98}} $(\pm {.24})$  & {\textbf{94.90}} $(\pm {.16})$ 
    & {\textbf{85.15}} $(\pm {.08})$ & {\textbf{68.85}} $(\pm {.16})$ 
    & {\textbf{25.82}} $(\pm {.45})$ & {\textbf{44.61}} $(\pm {.11})$ 
    & {\textbf{66.92}} $(\pm {.04})$  \\
    
    \bottomrule
  \end{tabular}
}
  \caption{Acc@1 of zero-shot image classification with CLIP-ViT-B/16 backbone on cross-dataset generalization over 3 random seeds. Best performances are in \textbf{bold}.
  }
  \label{tab:cross_dataset_more_seeds}
\end{table*}

From Table~\ref{tab:cross_dataset_more_seeds}, we see that our conclusion from Sec~\ref{sec:main_results_cross} remains valid. Specifically, TPS outperforms TPT~\cite{shu2022tpt} by $> 2\%$ on average. Similarly to Sec~\ref{sec:natural_dist_more_seeds}, we observe that taking the mean of the TPT-tuned and advanced off-the-shelf prototypes increases performance by only $0.5\%$ on average over TPT, demonstrating TPT's inflexibility in utilizing these more robust class representations.

\subsection{Context-Dependent Visual Reasoning}
\label{sec:bongard_hoi_more_seeds}

\begin{table}[t]
\centering
\resizebox{\columnwidth}{!}{
\small
  \centering
  \resizebox{\linewidth}{!}{%
  \begin{tabular}{l|*{5}{c}}
    \toprule
    \textbf{Method} & \textbf{seen act.,} & \textbf{unseen act.,} & \textbf{seen act.,} & \textbf{unseen act.,} & \textbf{Average} \\
    & \textbf{seen obj.,} & \textbf{seen obj.,} & \textbf{unseen obj.,} & \textbf{unseen obj.,} & \\
    \midrule

    TPT (reprod.)
    &  {65.81 ($\pm .12$)} & {69.15 ($\pm .10$)} & {65.69 ($\pm .01$)} & {\textbf{66.87} ($\pm .03$)} & {66.88 ($\pm .04$)} \\

    \textbf{Ours} (Shift)
    &  {\textbf{66.67 ($\pm 0.68$)}} & {\textbf{70.31 ($\pm 1.67$)}} & {\textbf{66.00 ($\pm 1.38$)}} & {66.67 ($\pm 0.40$)} & {\textbf{67.41 ($\pm .98$)}} \\
    
    \bottomrule
  \end{tabular}
}
}
  \caption{Acc on the Bongard-HOI benchmark with CLIP-ResNet-50 backbone over 3 random seeds. Best performances are in \textbf{bold}.}
  \label{tab:bongard_hoi_multiple_seeds}
\end{table}
From Table~\ref{tab:bongard_hoi_multiple_seeds}, we see that our conclusion from Sec~\ref{sec:main_results_bongard} remains valid. Specifically, TPS outperforms TPT~\cite{shu2022tpt} by $> 0.5\%$ on average.

\begin{table}[t]
  \centering

  \resizebox{\columnwidth}{!}{%
  \begin{tabular}{l|*{5}{C}}
    \toprule
    \textbf{Method}
    & \textbf{ImageNet} & \textbf{ImageNet-R} & \textbf{ImageNet-Sketch}  
    & \textbf{Cross-Dataset Average} & \textbf{Needs Support Set} \\

    \midrule
    SuS-X-SD-C~\cite{udandarao2022sus-x} 
    & {61.65} & {61.69} & {35.88} & {60.49} & {\ding{51}} \\

    SuS-X-LC-P~\cite{udandarao2022sus-x} 
    & {61.80} & {61.62} & {36.25} & {60.64} & {\ding{51}} \\

    CALIP~\cite{guo2022calip} 
    & {60.57} & {-} & {-} & {59.34} & {\ding{55}} \\

    \midrule

    \textbf{Ours} (Shift~+~SuS-X descriptors)
    & {\textbf{63.52}} & {\textbf{63.66}} & {\textbf{37.66}} & {\textbf{61.47}} & {\ding{55}} \\

    \bottomrule
  \end{tabular}
}
  \caption{Acc@1 of zero-shot image classification with CLIP-ResNet-50 backbone on ImageNet and its OOD variants. Best performances are in \textbf{bold}.
  }
    \label{tab:training_free}
\end{table}

\section{Full Ablations}\label{sec:s3}

In Sec~\ref{sec:shift_on_diff_prototype_appendix}, we report full ablations on TPS on the effectiveness of feature-space shift on various prototypes. These results are comparable to those reported in Sec~\ref{sec:ablations}. In Sec~\ref{sec:per_class_v_shared_appendix}, we include additional ablations to observe the effect of learning a class-specific shift over a universal shift for all classes. In Sec~\ref{sec:prototype_variants_appendix}, we explore variants on prototype generation using the class-agnostic CLIP ImageNet context prompt templates~\cite{radford2021clip} and the class-specific descriptors generated using GPT-4~\cite{openai2023gpt4}. All these ablations are run over 3 random seeds.

\subsection{Effect of Shift on Different Prototypes}
\label{sec:shift_on_diff_prototype_appendix}

\begin{table*}[t]
\small
  \centering
  \resizebox{\linewidth}{!}{%
  \begin{tabular}{l|C|*{7}C}
    \toprule
    \textbf{Prompt Type} & \textbf{Setting}
    & \textbf{ImageNet} &  \textbf{ImageNet-A} & \textbf{ImageNet-V2}
    & \textbf{ImageNet-R} & \textbf{ImageNet-Sketch} 
    & \textbf{Average} & \textbf{OOD Average} \\
    \midrule

    \multirow{3}{*}{Vanilla} & Zero-Shot
    & {66.74} & {47.79} & {60.89}
    & {73.99} & {46.12} 
    & {59.10} & {57.20} \\

    & ~+~shift
    &  {68.81} $(\pm {.03})$  & {58.11} $(\pm {.16})$  & {63.51} $(\pm {.17})$ 
    &  {76.98} $(\pm {.05})$  & {48.11} $(\pm {.09})$  
    & {63.10} $(\pm {.08})$ & {61.68} $(\pm {.09})$  \\

    \midrule
    & $\mathbf{\Delta}$ 
    &  \cellcolor{Gray2}{\green{{~+~2.07}}}  & \cellcolor{Gray2}{\green{{~+~10.32}}}  & \cellcolor{Gray2}{\green{{~+~2.62}}}
    &  \cellcolor{Gray2}{\green{{~+~2.99}}}  & \cellcolor{Gray2}{\green{{~+~1.99}}} 
    &  \cellcolor{Gray2}{\green{{~+~4.00}}}  & \cellcolor{Gray2}{\green{{~+~4.48}}}  \\

    \midrule
    \midrule
    
    \multirow{3}{*}{CoOp~\cite{zhou2021coop}} & Zero-Shot
    & {71.51} & {49.71} & {64.20}
    & {75.21} & {47.99} 
    & {61.72} & {59.28} \\

    & ~+~shift
    & {73.76} $(\pm {.04})$  & {60.43} $(\pm {.12})$  & {66.84} $(\pm {.10})$ 
    &  {77.39} $(\pm {.05})$  & {49.08} $(\pm {.06})$  
    & {65.50} $(\pm {.02})$ & {63.44} $(\pm {.03})$  \\

    \midrule
    & $\mathbf{\Delta}$
    &  \cellcolor{Gray2}{\green{{~+~2.25}}}  & \cellcolor{Gray2}{\green{{~+~10.72}}}  & \cellcolor{Gray2}{\green{{~+~2.64}}} 
    &  \cellcolor{Gray2}{\green{{~+~2.18}}}  & \cellcolor{Gray2}{\green{{~+~1.09}}}  
    &  \cellcolor{Gray2}{\green{{~+~3.78}}}  & \cellcolor{Gray2}{\green{{~+~4.16}}}  \\

    \midrule
    \midrule
    
    \multirow{3}{*}{CLIP templates} & Zero-Shot
    &  {68.35}  & {49.95}  & {61.97} 
    &  {77.59}  & {48.21}  
    & {61.21} & {59.43}  \\

    & ~+~shift
    &  {70.39} $(\pm {.06})$  & {60.47} $(\pm {.07})$  & {64.66} $(\pm {.04})$ 
    &  {80.70} $(\pm {.04})$  & {50.38} $(\pm {.14})$  
    & {65.32} $(\pm {.03})$ & {64.05} $(\pm {.02})$  \\

    \midrule
    & $\mathbf{\Delta}$
    &  \cellcolor{Gray2}{\green{{~+~2.04}}}  & \cellcolor{Gray2}{\green{{~+~10.52}}}  & \cellcolor{Gray2}{\green{{~+~2.69}}} 
    &  \cellcolor{Gray2}{\green{{~+~3.11}}}  & \cellcolor{Gray2}{\green{{~+~2.17}}} 
    &  \cellcolor{Gray2}{\green{{~+~4.11}}}  & \cellcolor{Gray2}{\green{{~+~4.62}}}  \\

    \midrule
    \midrule

    \multirow{3}{*}{Descriptors} & Zero-Shot
    &  {68.52}  & {48.91}  & {61.78} 
    &  {74.81}  & {47.68}  
    & {60.34} & {58.29}  \\

    & ~+~shift
    &  {70.38} $(\pm {.03})$  & {59.21} $(\pm {.09})$  & {63.80} $(\pm {.07})$ 
    &  {77.49} $(\pm {.12})$  & {49.57} $(\pm {.06})$  
    & {64.09} $(\pm {.02})$ & {62.52} $(\pm {.03})$  \\

    \midrule
    & $\mathbf{\Delta}$
    &  \cellcolor{Gray2}{\green{{~+~1.86}}}  & \cellcolor{Gray2}{\green{{~+~10.30}}}  & \cellcolor{Gray2}{\green{{~+~2.02}}} 
    &  \cellcolor{Gray2}{\green{{~+~2.68}}}  & \cellcolor{Gray2}{\green{{~+~1.89}}}  
    &  \cellcolor{Gray2}{\green{{~+~3.75}}}  & \cellcolor{Gray2}{\green{{~+~4.23}}}  \\

    \midrule
    \midrule
    
    CLIP templates
    & Zero-Shot
    &  {69.54}  & {50.51}  & {63.01}
    &  {77.18}  & {48.84}  
    & {61.82} & {59.88} \\

    ~+~Descriptors 
    & ~+~shift
    &  {71.43} $(\pm {.06})$  & {60.78} $(\pm {.21})$  & {65.00} $(\pm {.09})$ 
    &  {80.06} $(\pm {.13})$  & {50.97} $(\pm {.09})$  
    & {65.65} $(\pm {.06})$ & {64.20} $(\pm {.08})$  \\

    \midrule
    & $\mathbf{\Delta}$
    &  \cellcolor{Gray2}{\green{~+~1.89}}  & \cellcolor{Gray2}{\green{{~+~10.27}}}  & \cellcolor{Gray2}{\green{{~+~1.99}}} 
    &  \cellcolor{Gray2}{\green{{~+~2.88}}}  & \cellcolor{Gray2}{\green{{~+~2.13}}} 
    &  \cellcolor{Gray2}{\green{{~+~3.83}}}  & \cellcolor{Gray2}{\green{{~+~4.32}}}  \\

    \bottomrule
  \end{tabular}
}
  \caption{Acc@1 for zero-shot and with feature-space shift with features initialized using different prototype generation techniques on ImageNet and its out-of-distribution variants. Results are over 3 random seeds.
  }
  \label{tab:zero_shot_vs_shift_natural_more_seeds}
\end{table*}
\begin{table*}[t]
  \centering
  \resizebox{\linewidth}{!}{%
  \begin{tabular}{l|c|c*{11}c}
    \toprule
    \textbf{Prompt Type} & \textbf{Setting}
    & \textbf{Flower102} &  \textbf{DTD} & \textbf{Pets} & \textbf{Cars}
    & \textbf{UCF101}   & \textbf{CalTech101} & \textbf{Food101}   & \textbf{SUN397}
    & \textbf{Aircraft} & \textbf{EuroSAT} & \textbf{Average} \\
    \midrule

    \multirow{3}{*}{Vanilla} & Zero-Shot
    & {67.28} & {44.44} & {87.98} & {65.24}
    & {65.08} & {92.98} & {83.80} & {62.55}
    & {23.70} & {41.42} & {63.45} \\

    & ~+~shift
    & {67.75} $(\pm {.10})$  & {45.69} $(\pm {.10})$  
    & {87.57} $(\pm {.10})$ &  {67.60} $(\pm {.23})$ 
    & {66.79} $(\pm {.21})$  & {93.79} $(\pm {.08})$ 
    & {84.62} $(\pm {.03})$ & {64.58} $(\pm {.03})$ 
    & {24.75} $(\pm {.39})$ & {41.35} $(\pm {.03})$ 
    & {64.45} $(\pm {.04})$  \\

    \midrule
    & $\mathbf{\Delta}$
    &  \cellcolor{Gray2}{\green{{~+~0.47}}}  & \cellcolor{Gray2}{\green{{~+~1.25}}}  & \cellcolor{Gray2}{\red{{~-~0.41}}} 
    &  \cellcolor{Gray2}{\green{{~+~2.36}}}  & \cellcolor{Gray2}{\green{{~+~1.71}}} 
    &  \cellcolor{Gray2}{\green{{~+~0.81}}}  & \cellcolor{Gray2}{\green{{~+~0.82}}} & \cellcolor{Gray2}{\green{{~+~2.03}}} & \cellcolor{Gray2}{\green{{~+~1.05}}} & \cellcolor{Gray2}{\red{{~-~0.07}}} & \cellcolor{Gray2}{\green{{~+~1.00}}}  \\

    \midrule
    \midrule
    
    \multirow{3}{*}{CLIP templates} & Zero-Shot
    &  {65.57}  & {44.86}  & {88.25} & {66.19}
    &  {67.46}  & {93.67}  & {83.77} & {65.78}
    &  {23.64}  & {47.74}  & {64.69}  \\

    & ~+~shift
    & {66.41} $(\pm {.05})$  & {45.61} $(\pm {.19})$  
    & {87.99} $(\pm {.10})$ &  {68.66} $(\pm {.31})$ 
    & {68.02} $(\pm {.11})$  & {93.85} $(\pm {.14})$ 
    & {84.54} $(\pm {.08})$ & {67.19} $(\pm {.05})$ 
    & {24.66} $(\pm {.13})$ & {48.28} $(\pm {.20})$ 
    & {65.52} $(\pm {.05})$  \\

    \midrule
    & $\mathbf{\Delta}$
    &  \cellcolor{Gray2}{\green{{~+~0.84}}}  & \cellcolor{Gray2}{\green{{~+~0.75}}}  & \cellcolor{Gray2}{\red{{~-~0.26}}} 
    &  \cellcolor{Gray2}{\green{{~+~2.47}}}  & \cellcolor{Gray2}{\green{{~+~0.56}}} 
    &  \cellcolor{Gray2}{\green{{~+~0.18}}}  & \cellcolor{Gray2}{\green{{~+~0.77}}} & \cellcolor{Gray2}{\green{{~+~1.41}}} & \cellcolor{Gray2}{\green{{~+~1.02}}} & \cellcolor{Gray2}{\green{{~+~0.54}}} & \cellcolor{Gray2}{\green{{~+~0.83}}}  \\

    \midrule
    \midrule

    \multirow{3}{*}{Descriptors} & Zero-Shot
    & {71.13} & {52.72} & {86.75} & {65.15}
    & {70.53} & {94.08} & {84.12} & {67.10}
    & {25.26} & {43.31} & {66.02} \\

    & ~+~shift
    & {71.69} $(\pm {.15})$  & {53.80} $(\pm {.21})$  
    & {87.82} $(\pm {.19})$ &  {67.00} $(\pm {.14})$ 
    & {71.18} $(\pm {.15})$  & {94.56} $(\pm {.08})$ 
    & {84.78} $(\pm {.05})$ & {68.25} $(\pm {.18})$ 
    & {26.27} $(\pm {.09})$ & {42.11} $(\pm {.18})$ 
    & {66.75} $(\pm {.06})$  \\

    \midrule
    & $\mathbf{\Delta}$
    &  \cellcolor{Gray2}{\green{{~+~0.56}}}  & \cellcolor{Gray2}{\green{{~+~1.08}}}  & \cellcolor{Gray2}{\green{{~+~1.07}}} 
    &  \cellcolor{Gray2}{\green{{~+~1.85}}}  & \cellcolor{Gray2}{\green{{~+~0.65}}} 
    &  \cellcolor{Gray2}{\green{{~+~0.48}}}  & \cellcolor{Gray2}{\green{{~+~0.66}}} & \cellcolor{Gray2}{\green{{~+~1.15}}} & \cellcolor{Gray2}{\green{{~+~1.01}}} & \cellcolor{Gray2}{\red{{~-~1.20}}} & \cellcolor{Gray2}{\green{{~+~0.73}}}  \\

    \midrule
    \midrule
    
    CLIP templates
    & Zero-Shot
    & {70.52} & {49.94} & {87.22} & {66.48}
    & {70.24} & {94.12} & {84.47} & {67.55}
    & {24.69} & {44.14} & {65.94} \\

    ~+~Descriptors 
    & ~+~shift
    & {71.47} $(\pm {.12})$  & {51.00} $(\pm {.47})$  
    & {87.45} $(\pm {.09})$ &  {68.99} $(\pm {.10})$ 
    & {70.98} $(\pm {.24})$  & {94.90} $(\pm {.16})$ 
    & {85.15} $(\pm {.08})$ & {68.85} $(\pm {.16})$ 
    & {25.82} $(\pm {.45})$ & {44.61} $(\pm {.11})$ 
    & {66.92} $(\pm {.04})$  \\

    \midrule
    & $\mathbf{\Delta}$
    &  \cellcolor{Gray2}{\green{{~+~0.95}}}  & \cellcolor{Gray2}{\green{{~+~1.06}}}  & \cellcolor{Gray2}{\green{{~+~0.23}}} 
    &  \cellcolor{Gray2}{\green{{~+~2.51}}}  & \cellcolor{Gray2}{\green{{~+~0.74}}} 
    &  \cellcolor{Gray2}{\green{{~+~0.78}}}  & \cellcolor{Gray2}{\green{{~+~0.68}}} & \cellcolor{Gray2}{\green{{~+~1.30}}} & \cellcolor{Gray2}{\green{{~+~1.13}}} & \cellcolor{Gray2}{\green{{~+~0.47}}} & \cellcolor{Gray2}{\green{{~+~0.98}}}  \\
    
    \bottomrule
  \end{tabular}
}
  \caption{Acc@1 for zero-shot and with feature-space shift with features initialized using different prototype generation techniques on cross-domain generalization datasets. Results are over 3 random seeds.
  }
  \label{tab:zero_shot_vs_shift_cross_more_seeds}
\end{table*}

Full comparisons between zero-shot and feature-shifted performance on all natural distribution shift and cross-domain generalization benchmark datasets over 3 random seeds are in Tables~\ref{tab:zero_shot_vs_shift_natural_more_seeds} and ~\ref{tab:zero_shot_vs_shift_cross_more_seeds}, respectively. We demonstrate that our conclusion from Sec~\ref{sec:effect_of_shift} stills holds -- that learning a small perturbation in the feature space results in performance gains of $>4\%$ and up to $1\%$ on average across natural distribution shift and cross-domain generalization tasks regardless of what prototypes are used.

\subsection{Effect of Per-Class vs. Shared Shift}
\label{sec:per_class_v_shared_appendix}
\begin{table*}[t]
\small
  \centering
  \resizebox{0.98\linewidth}{!}{%
  \begin{tabular}{l|*{7}C}
    \toprule
    \textbf{Method}
    & \textbf{ImageNet}          &  \textbf{ImageNet-A} 
    & \textbf{ImageNet-V2}       & \textbf{ImageNet-R}
    & \textbf{ImageNet-Sketch}   
    & \textbf{Average} & \textbf{OOD Average} \\
    \midrule

    Shared
    &  {71.23} $(\pm {.02})$  & {56.57} $(\pm {.19})$  & {64.98} $(\pm {.03})$ 
    &  {79.31} $(\pm {.03})$  & {50.80} $(\pm {.06})$  
    & {64.58} $(\pm {.04})$ & {62.92} $(\pm {.06})$  \\
    Per class
    &  {\textbf{71.43}} $(\pm {.06})$  & {\textbf{60.78}} $(\pm {.21})$  & {\textbf{65.00}} $(\pm {.09})$ 
    &  {\textbf{80.06}} $(\pm {.13})$  & {\textbf{50.97}} $(\pm {.09})$  
    & {\textbf{65.65}} $(\pm {.06})$ & {\textbf{64.20}} $(\pm {.08})$  \\
    
    \bottomrule
  \end{tabular}
    }
  \caption{Acc@1 for learning a shared vs. per-class shift on top of different prototypes over 3 random seeds. Best performances are in \textbf{bold}.
  }
  \label{tab:shift_natural_more_seeds}
\end{table*}
\begin{table*}[t]
\small
  \centering
  \resizebox{0.98\linewidth}{!}{%
  \begin{tabular}{l|*{11}c}
    \toprule
    \textbf{Method}
    & \textbf{Flower102} &  \textbf{DTD} & \textbf{Pets} & Cars
    & \textbf{UCF101}   & \textbf{CalTech101} & \textbf{Food101}   & \textbf{SUN397}
    & \textbf{Aircraft} & \textbf{EuroSAT} & \textbf{Average} \\
    \midrule

    Shared 
    &  {71.36} $(\pm {.12})$  & {50.49} $(\pm {.12})$  & {\textbf{87.46}} $(\pm {.12})$ 
    &  {67.33} $(\pm {.06})$  & {70.77} $(\pm {.12})$  
    & {94.35} $(\pm {.06})$ & {84.82} $(\pm {.01})$ & {68.12} $(\pm {.04})$ & {25.27} $(\pm {.02})$ & {\textbf{44.67}} $(\pm {.06})$ & {66.47} $(\pm {.03})$ \\
    Per-class
    & {\textbf{71.47}} $(\pm {.12})$  & {\textbf{51.00}} $(\pm {.47})$  
    & {87.45} $(\pm {.09})$ &  {\textbf{68.99}} $(\pm {.10})$ 
    & {\textbf{70.98}} $(\pm {.24})$  & {\textbf{94.90}} $(\pm {.16})$ 
    & {\textbf{85.15}} $(\pm {.08})$ & {\textbf{68.85}} $(\pm {.16})$ 
    & {\textbf{25.82}} $(\pm {.45})$ & {44.61} $(\pm {.11})$ 
    & {\textbf{66.92}} $(\pm {.04})$  \\
    
    \bottomrule
  \end{tabular}
    }
  \caption{Acc@1 for learning a shared vs. per-class shift on top of different prototypes over 3 random seeds. Best performances are in \textbf{bold}.
  }
  \label{tab:shift_cross_more_seeds}
\end{table*}

Test-time prompt tuning methods involve tuning a prompt that is shared across all classes in a dataset. Given that the tuneable prompt tokens form a portion of the text encoder input, these full prompts are then mapped to the embedding space with the encoder's learned complex feature-space mapping. This results in non-linear perturbations from the original class prototypes. However, for our method, tuning shift parameters that are shared for all class prototypes in the feature-space means that the relative distance between class prototypes will remain constant before and after test-time shift tuning, limiting the expressive capability of the learned shift. Rather, we believe that each class prototype should be modulated by slightly different magnitudes and/or directions to provide more degrees of freedom in capturing the class-level distribution shifts in addition to the dataset-level shifts present in a domain gap.

Table~\ref{tab:shift_natural_more_seeds} shows that, on average, learning a per-class shift increases performance by $> 1.2\%$ regardless of which prototypes are used. Moreover, we see that Table~\ref{tab:shift_cross_more_seeds} demonstrates that, on average, learning a per-class shift increases performance by around $0.5\%$ on average over different prototype settings. This demonstrates that learning per-class shifts allows the model to capture both dataset-level and class-level distribution shifts in a domain gap. 

\subsection{Prototype Variants}
\label{sec:prototype_variants_appendix}
\begin{table*}[t]
\small
  \centering
  \resizebox{0.98\linewidth}{!}{%
  \begin{tabular}{l|C|*{7}C}
    \toprule
    \textbf{Prompt Type(s)} & \textbf{Pooling Method} 
    & \textbf{ImageNet} &  \textbf{ImageNet-A} & \textbf{ImageNet-V2}
    & \textbf{ImageNet-R} & \textbf{ImageNet-Sketch} 
    & \textbf{Average} & \textbf{OOD Average} \\
    \midrule

    & & \multicolumn{7}{c}{\textit{\underline{Zero-Shot}}}  \\
    Vanilla prompt & N/A
    & {66.74} & {47.79} & {60.89}
    & {73.99} & {46.12} 
    & {59.10} & {57.20} \\
    CLIP templates~+~Descriptors & Macro
    &  {68.73}  & {50.32}  & {62.31} 
    &  {\textbf{77.67}}  & {48.56}  
    & {61.52} & {59.72}  \\
    CLIP templates~+~Descriptors & Micro
    &  {\textbf{69.54}}  & {50.51}  & {\textbf{63.01}} 
    &  {77.18}  & {48.84}  
    & {\textbf{61.82}} & {\textbf{59.88}}  \\
    CLIP templates~$\times$~Descriptors & Macro 
    &  {69.03}  & {\textbf{50.73}}  & {62.22} 
    &  {76.91}  & {\textbf{49.07}}  
    & {61.59} & {59.73}  \\

    \midrule

    & & \multicolumn{7}{c}{\textit{\underline{With Shift}}}  \\
    Vanilla prompt & N/A 
    &  {68.81} $(\pm {.03})$  & {58.11} $(\pm {.16})$  & {63.51} $(\pm {.17})$ 
    &  {76.98} $(\pm {.05})$  & {48.11} $(\pm {.09})$  
    & {63.10} $(\pm {.08})$ & {61.68} $(\pm {.09})$  \\

    CLIP templates~+~Descriptors & Macro 
    &  {70.75} $(\pm {.08})$  & {\textbf{60.86}} $(\pm {.09})$  & {64.95} $(\pm {.11})$ 
    &  {\textbf{80.84}} $(\pm {.03})$  & {50.70} $(\pm {.11})$  
    & {65.62} $(\pm {.02})$ & {\textbf{64.34}} $(\pm {.02})$  \\
    
    CLIP templates~+~Descriptors & Micro
    &  {\textbf{71.43}} $(\pm {.06})$  & {60.78} $(\pm {.21})$  & {\textbf{65.00}} $(\pm {.09})$ 
    &  {80.06} $(\pm {.13})$  & {50.97} $(\pm {.09})$  
    & {\textbf{65.65}} $(\pm {.06})$ & {64.20} $(\pm {.08})$  \\
    
    CLIP templates~$\times$~Descriptors & Macro 
    &  {70.82} $(\pm {.02})$  & {60.42} $(\pm {.06})$  & {64.50} $(\pm {.05})$ 
    &  {79.53} $(\pm {.09})$  & {\textbf{51.13}} $(\pm {.02})$  
    & {65.28} $(\pm {.01})$ & {63.89} $(\pm {.02})$  \\
    \bottomrule
  \end{tabular}
}
  \caption{Acc@1 for different variants of prototype generation, i.e. ways of combining templates and descriptors, on natural distribution shifts, over 3 random seeds. Best performances for each setting are in \textbf{bold}.
  }
  \label{tab:knowledge_injection_variants_imagenet}
\end{table*}
\begin{table*}[!htb]
\small
  \centering
  \resizebox{0.98\linewidth}{!}{%
  \begin{tabular}{l|c|*{11}c}
    \toprule
   \textbf{ Prompt Type(s)} & \textbf{Pooling Method}
    & \textbf{Flower102} &  \textbf{DTD} & \textbf{Pets} & \textbf{Cars}
    & \textbf{UCF101}   & \textbf{CalTech101} & \textbf{Food101}   & \textbf{SUN397}
    & \textbf{Aircraft} & \textbf{EuroSAT} & \textbf{Average} \\
    \midrule

    & & \multicolumn{11}{c}{\textit{\underline{Zero-Shot}}}  \\
    Vanilla prompt & N/A
    & {67.28} & {44.44} & {87.98} & {65.24}
    & {65.08} & {92.98} & {83.80} & {62.55}
    & {23.70} & {41.42} & {63.45} \\
    CLIP templates~+~Descriptors & Macro
    &  {66.91}  & {45.86}  & {\textbf{88.33}} & {66.46}
    &  {68.12}  & {93.83}  & {83.97} & {66.34}
    &  {24.03}  & {46.62}  & {65.05}  \\
    CLIP templates~+~Descriptors & Micro
    &  {70.52}  & {49.94}  & {87.22} & {\textbf{66.48}}
    &  {70.24}  & {94.12}  & {\textbf{84.47}} & {67.55}
    &  {24.69}  & {44.14}  & {65.94}  \\
    CLIP templates~$\times$~Descriptors & Macro
    &  {\textbf{72.03}}  & {\textbf{50.83}}  & {86.21} & {66.12}
    &  {\textbf{70.90}}  & {\textbf{94.16}}  & {83.73} & {\textbf{67.98}}
    &  {\textbf{25.53}}  & {\textbf{47.19}}  & {\textbf{66.47}}  \\

    \midrule
    & & \multicolumn{11}{c}{\textit{\underline{With Shift}}}  \\
    Vanilla prompt & N/A 
    & {67.75} $(\pm {.10})$  & {45.69} $(\pm {.10})$  
    & {87.57} $(\pm {.10})$ &  {67.60} $(\pm {.23})$ 
    & {66.79} $(\pm {.21})$  & {93.79} $(\pm {.08})$ 
    & {84.62} $(\pm {.03})$ & {64.58} $(\pm {.03})$ 
    & {24.75} $(\pm {.39})$ & {41.35} $(\pm {.03})$ 
    & {64.45} $(\pm {.04})$  \\
    CLIP templates~+~Descriptors & Macro
    &  {67.52} $(\pm {.27})$  & {46.43} $(\pm {.28})$  & {\textbf{88.00}} $(\pm {.13})$ 
    &  {\textbf{69.04}} $(\pm {.16})$  & {68.67} $(\pm {.18})$  
    & {94.16} $(\pm {.18})$ & {84.77} $(\pm {.04})$  & {67.70} $(\pm {.08})$ & {24.79} $(\pm {.30})$  & {\textbf{47.09}} $(\pm {.19})$ & {65.82} $(\pm {.06})$ \\
    CLIP templates~+~Descriptors & Micro
    & {71.47} $(\pm {.12})$  & {51.00} $(\pm {.47})$  
    & {87.45} $(\pm {.09})$ &  {68.99} $(\pm {.10})$ 
    & {70.98} $(\pm {.24})$  & {\textbf{94.90}} $(\pm {.16})$ 
    & {\textbf{85.15}} $(\pm {.08})$ & {68.85} $(\pm {.16})$ 
    & {25.82} $(\pm {.45})$ & {44.61} $(\pm {.11})$ 
    & {66.92} $(\pm {.04})$  \\
    CLIP templates~$\times$~Descriptors & Macro 
    &  {\textbf{72.53}} $(\pm {.12})$  & {\textbf{52.56}} $(\pm {.09})$  & {86.15} $(\pm {.05})$ 
    &  {68.89} $(\pm {.07})$  & {\textbf{71.44}} $(\pm {.20})$  
    & {94.43} $(\pm {.06})$ & {84.44} $(\pm {.08})$  & {\textbf{69.04}} $(\pm {.02})$ & {\textbf{26.51}} $(\pm {.26})$  & {45.65} $(\pm {.15})$ & {\textbf{67.16}} $(\pm {.03})$ \\
    \bottomrule
  \end{tabular}
}
  \caption{Acc@1 for different variants of knowledge injection, i.e. ways of combining templates and descriptors, over 3 random seeds on cross-dataset generalization tasks. Best performances in each setting are in \textbf{bold}.
  }
\label{tab:knowledge_injection_variants_cross}
\end{table*}

We explore different methods for creating class prototypes. Specifically, we experiment with different forms of aggregating the text encoded with the 80 ImageNet context prompts from CLIP~\cite{radford2021clip} and our LLM-generated descriptors. The CLIP ImageNet templates are class-agnostic and add image-level characteristics whereas the descriptors are class-specific and add class-level semantic information.

Tables~\ref{tab:knowledge_injection_variants_imagenet} and ~\ref{tab:knowledge_injection_variants_cross} compare three variants of pooling these CLIP templated embeddings and descriptor embeddings to obtain a single class prototype. Similarly to the conclusion of Sec~\ref{sec:effect_of_shift}, we observe that in general, the gains observed using more advanced prototypes in the zero-shot setting almost directly translate to the test-time adaptation setting with shifting. In Sec~\ref{sec:experimental_results}, we present the results of our method using prototypes that are a micro average of the CLIP templates and LLM-generated descriptors.

\section{Comparison to Training-Free Methods}\label{sec:s4}

In a similar spirit to zero-shot test-time adaptation, training-free methods perform domain adaptation without tuning any parameters. We compare our method to state-of-the-art training-free methods in Table~\ref{tab:training_free}. We show that our method, TPS, when using the same GPT-3-generated~\cite{gpt3} text prompts from the official SuS-X~\cite{udandarao2022sus-x} code, out-performs both CALIP~\cite{guo2022calip} and SuS-X~\cite{udandarao2022sus-x} without any additional image support set constructed with Stable Diffusion~\cite{Rombach_2022_CVPR} or LAION-5B~\cite{laion5b}. This demonstrates how a simple feature-space shift is more effective than complex training-free methods. For fair comparison, we compare TPS with SuS-X results with fixed hyperparameter settings as ours are not tuned per dataset and use the same CLIP-ResNet50 backbone.

% \newpage
% \section{Theoretical Analysis of Feature Space Shift}\label{sec:s4}
% \input{sec/gradient}

\section{Research Impact and Limitations} \label{sec:s5}
We propose TPS, a framework that can be used to easily and effectively improve zero-shot generalization of VLMs. Given the large-scale training of foundation VLMs, we believe it is important to understand different ways to better leverage the resulting rich multi-modal contrastive representation spaces in parameter- and runtime-constrained settings. We propose to learn a slight perturbation to the class prototypes to maintain the overall representation quality of the pre-trained embedding space while learning a better alignment to the OOD target dataset. We hope that this framework can inspire future work to explore other tasks where learning directly in the feature space can be an efficient alternative to more complex tuning approaches. 

Our work builds on the CLIP~\cite{radford2021clip} representation space and uses GPT-4~\cite{openai2023gpt4} to generate class descriptors to create more advanced class prototypes. Thus, our model has the potential to magnify the biases of both these models. Future studies may explore how to best leverage these models' capabilities without promoting its biases.

\end{document}